\documentclass[
]{ceurart}

\usepackage{listings}
\usepackage{tikz}

\usepackage{algorithm}
\usepackage{algorithmic}
\usepackage{amsthm}
\usepackage{amsmath}
\usepackage{amsfonts}
\usepackage{amssymb}
\theoremstyle{plain}
\newtheorem{theorem}{Theorem}[section]

\theoremstyle{definition}
\newtheorem{definition}[theorem]{Definition}

\theoremstyle{remark}

\begin{document}

\copyrightyear{2024}
\copyrightclause{Copyright for this paper by its authors.
  Use permitted under Creative Commons License Attribution 4.0
  International (CC BY 4.0).}

\conference{xAI-2024: The 2nd World Conference on eXplainable Artificial Intelligence, July 17--19, 2024, Valletta, Malta}

\title{AnyCBMs: How to Turn Any Black Box into a Concept Bottleneck Model}

\tnotemark[1]

\author[1]{Gabriele Dominici}[%
orcid=0009-0009-1955-0778,
email=gabriele.dominici@usi.ch,
url=https://gabriele-dominici.github.io/
]
\cormark[1]
\fnmark[1]
\address[1]{Università della Svizzera Italiana,
  Lugano, Switzerland}

\author[1]{Pietro Barbiero}[%
orcid=,
email=pietro.barbiero@usi.ch,
]
\fnmark[1]

\author[2]{Francesco Giannini}[%
orcid=,
email=francesco.giannini@unisi.it,
]
\address[2]{Scuola Normale Superiore, Pisa, Italy}

\author[1]{Martin Gjoreski}[%
orcid=,
email=martin.gjoreski@usi.ch,
]

\author[1]{Marc Langeinrich}[%
orcid=,
email=marc.langeinrich@usi.ch,
]

\cortext[1]{Corresponding author.}
\fntext[1]{These authors contributed equally.}

\begin{abstract}
Interpretable deep learning aims at developing neural architectures whose decision-making processes could be understood by their users. Among these techniqes, Concept Bottleneck Models enhance the interpretability of neural networks by integrating a layer of human-understandable concepts. These models, however, necessitate training a new model from the beginning, consuming significant resources and failing to utilize already trained large models. To address this issue, we introduce ``AnyCBM'', a method that transforms any existing trained model into a Concept Bottleneck Model with minimal impact on computational resources. We provide both theoretical and experimental insights showing the effectiveness of AnyCBMs in terms of classification performances and effectivenss of concept-based interventions on downstream tasks.
\end{abstract}


\maketitle

\section{Introduction}
 Numerous national and international regulatory frameworks underscore the transformative potential of artificial intelligence (AI). However, they also warn of the inherent risks associated with such powerful technology, emphasizing the importance of careful monitoring and strict protections. For instance, the recent AI Act \cite{madiega2021artificial} aims to implement detailed regulations for AI systems, ensuring their safety, transparency, and accountability. Similarly, in the US, the federal government issued an executive order that proposes principles for trustworthy AI. Hence, interpretable AI has become a crucial aspect of modern machine learning to address concerns over the opaque nature of deep learning (DL) models~\citep{bussone2015role, rudin2019stop}. The quest for transparency has been driven by the need to understand the decision-making processes of AI systems, particularly in critical areas where ethical~\citep{duran2021afraid} and legal~\citep{lo2020ethical} implications of these systems' decisions are significant.

Concept Bottleneck Models (CBMs)~\citep{koh2020concept} are a family of differentiable models aiming to increase DL interpretability~\citep{ghorbani2019interpretation}. These models map input data (e.g., pixel intensities) to human-understandable concepts (e.g., shapes, colors), and then use these concepts to predict labels of a downstream classification task. However, existing CBMs necessitate training a new model from the beginning even in settings where trained or fine-tuned models already exist. In these scenarios, current CBM architectures would consume significant resources in re-training or fine-tuning again possibly large models. As a result, this limitation restricts CBMs' ability to be adopted in new domains.

\begin{figure}
    \centering
    \includegraphics[width=0.9\textwidth]{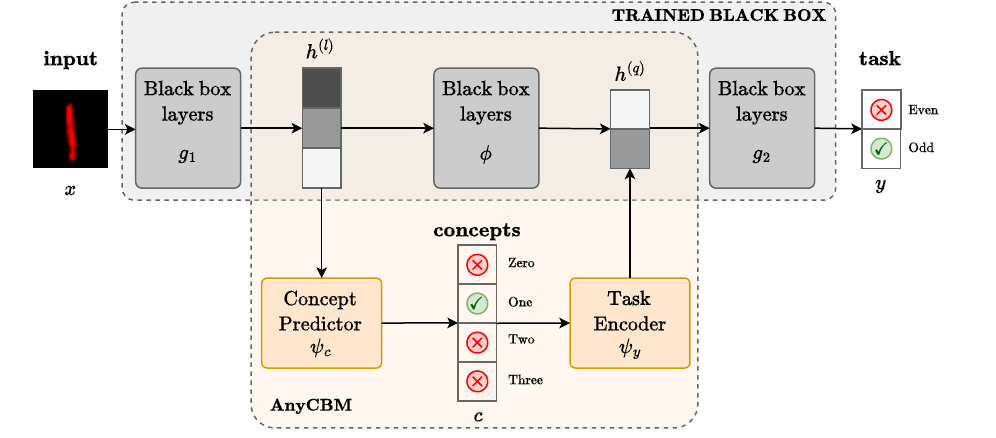}
    \caption{Any Concept Bottleneck Models (AnyCBMs) transform any black box neural architecture into an interpretable CBM mapping black box embeddings into a set of supervised concepts and then mapping the predicted concepts back to black box embeddings. This allows AnyCBMs to be applied to any layer of a trained black box and to perform concept-based interventions as in standard CBMs.}
    \label{fig:abstract}
\end{figure}

To bridge this gap, we introduce Any Concept Bottleneck Models (AnyCBMs, Figure~\ref{fig:abstract}), a method to transform any black-box neural architecture into an interpretable CBM. The key innovation of AnyCBMs lies in a neural model mapping black-box embeddings into a set of supervised concepts and then mapping the predicted concepts back to black-box embeddings. This allows AnyCBMs to be applied to any layer of a trained black box and to perform concept-based interventions as in standard CBMs. Our results demonstrate that AnyCBMs match black-box performance in terms of classification accuracy in downstream tasks and CBM performance with respect to concept accuracy. In addition,  AnyCBM could steer the behaviour of a black-box model acting on human-understandable concepts as effectively as in CBMs.

\section{Background}
Concept-based models $f: C \rightarrow Y$ learn a map from a concept space $C$ to a task space $Y$~\cite{yeh2020completeness}. If concepts are semantically meaningful, then humans can interpret this mapping by tracing back predictions to the most relevant concepts~\cite{ghorbani2019interpretation}. When the features of the input space are hard for humans to reason about (such as pixel intensities), concept-based models work on the output of a concept-encoder mapping $g: X \rightarrow C$ from the input space $X$ to the concept space $C$~\cite{ghorbani2019towards}. These architectures are known as Concept Bottleneck Models (CBMs)~\citep{koh2020concept}.
In general, training a CBM model may require a dataset 
where each sample consists of input features $x\in X \subseteq \mathbb{R}^n$ (e.g., an image's pixels), $k$ ground truth concepts $c\in  C \subseteq \{0, 1\}^k$ (i.e., a binary vector with concept annotations, when available) and $o$ task labels $y \in  Y \subseteq \{0, 1\}^o$ (e.g., an image's classes). 
During training, a CBM is encouraged to align its predictions to task labels i.e., $y \approx \hat{y}=f(g(x))$. Similarly, a concept predictor can be supervised when concept labels are available i.e., $c \approx \hat{c} = g(x)$. We indicate concept and task predictions as $\hat{c}_i=(g(x))_i$ and $\hat{y}_j=(f(\hat{c}))_j$ respectively. 
When concept labels are not available, they can still be extracted in with unsupervised techniques~\cite{ghorbani2019towards,magister2021gcexplainer,oikarinen2023labelfree}, which make CBMs applicable to a wide range of applications.

\section{AnyCBM: Turning Black Boxes into Concept Bottleneck Models}
AnyCBM (Figure~\ref{fig:abstract}) is a method designed to convert any opaque neural network architecture into a Concept Bottleneck Model (CBM) that is interpretable. The fundamental innovation of AnyCBMs involves the use of an external model that processes embeddings from a trained black box model. These embeddings, denoted as $h^{(l)} \in H^{(l)} \subseteq \mathbb{R}^{l}$, are encoded into a set of supervised concepts $c \in C$. Subsequently, these concepts are mapped back into embeddings $h^{(q)} \in H^{(q)} \subseteq \mathbb{R}^{q}$. This process allows for the embedding space of the black box model to be translated into a more understandable and interpretable form, where each concept represents a meaningful feature or characteristic that explains the decision-making process of the neural network. The following definition formalizes AnyCBMs.

\begin{definition}[AnyCBM]
Given a black box model $\phi: H^l \to H^q$ and a set of concepts $C$, a AnyCBM is a tuple of models $(\psi_c,\psi_y)$ such that, the following diagram commutes:
\begin{align*}
& \begin{tikzpicture}
  \node (h1) at (0,0) {$H^{(l)}$};
  \node (h2) at (4,0) {$H^{(q)}$};
  \node (c) at (2,-2) {$C$};
  \draw[->] (h1) -- (h2) node [midway,above]{$\phi$};
  \draw[->] (h1) -- (c) node [midway,left]{$\psi_c$};
  \draw[->] (c) -- (h2) node [midway,right]{$\psi_y$};
\end{tikzpicture}
\end{align*} 
\end{definition}
More specifically, the concept predictor \( \psi_c: H^{(l)} \to C \) encodes black box embeddings into concepts, and the task encoder \( \psi_y: C \to H^{(q)} \) maps concepts back into black box embeddings.
In practice, the commutative diagram describes how the interpretable mapping through $C$ via $\psi_c$ and $\psi_y$ should be consistent with the direct transformation of the black box $\phi$. Also properties and capabilities of AnyCBMs can directly be derived from the commutative diagram as it constraints the relationships among the transformations $\psi_c$, $\psi_y$, and $\phi$. 


In the following we present two practical case studies.

\paragraph{Case 1: \( \phi \) is the identity function on \( H \)}
When \( \phi \) is the identity function, \( \phi(h^{(l)}) = h^{(l)} \) for all \( h^{(l)} \in H^{(l)} \), and \( H^{(l)} = H^{(q)} \). The diagram simplifies, and we have:
\[
\psi_y \circ \psi_c = \text{id}_H
\]
\begin{theorem}
    If \( \phi \) is the identity function on \( H \), then \( \psi_y \) is injective:
    \begin{equation}
        \phi = \text{id}_H \implies \psi_y: C \hookrightarrow H^{(q)}
    \end{equation}
\end{theorem}

\begin{proof}
Assume \( \psi_y(c_1) = \psi_y(c_2) \). Since \( \psi_c \) is surjective, there exist \( h_1, h_2 \in H^{(l)} \) such that \( \psi_c(h_1) = c_1 \) and \( \psi_c(h_2) = c_2 \). Then,
\[
h_1 = \psi_y(\psi_c(h_1)) = \psi_y(c_1) = \psi_y(c_2) = \psi_y(\psi_c(h_2)) = h_2.
\]
Thus, \( c_1 = c_2 \), proving that \( \psi_y \) is injective.
\end{proof}
\textbf{Significance:} This property implies that \( \psi_y \) can uniquely reconstruct elements of \( H^{(l)} \) from \( C \), despite \( \psi_c \) not being injective. For example, if \( \psi_c \) represents a lossy compression, then \( \psi_y \) could be an error-correcting decoding where no information is lost despite compression.

\paragraph{Case 2: independent training}
In many practical cases, concept predictors and task encoders are independently trained to reduce concept leakage~\citep{mahinpei2021promises}. In this common setting, we can prove another property of AnyCBMs task encoders.

\begin{theorem}
If $\psi_c$ and $\psi_y$ are independently trained and \( \phi \) is a multi-layer neural network, then \( \psi_y \) cannot be surjective.
\end{theorem}
\begin{proof}
Assume for contradiction that \( \psi_y \) is surjective. The surjectivity of \( \psi_y \) would require that every point in \( H^{(q)} \) is the image of some point in \( C \). Given the independent training, the domain of $\psi_y$ is finite, specifically \( 2^k \). Since \( H^{(q)} \subseteq \mathbb{R}^q \), the mapping \( \psi_y: C \rightarrow H^{(q)} \) must pull from a set with finite cardinality \( 2^k \) to cover $\mathbb{R}^q$ which is a contradiction. Hence, \( \psi_y \) cannot be surjective.
\end{proof}

\textbf{Significance:} This theorem indicates that the surjectivity of \( \psi_y \) depends on the way we train the concept bottleneck. This means that, under independent training, AnyCBMs are not invertible, even when $\phi$ represents an invertible transformation.


\section{Experiments}
Our experiments aim to answer the following questions:
\begin{itemize}
    \item How is AnyCBMs classification performance on concepts and downstream tasks compared to standard CBMs and black boxes?
    \item How effective are concept interventions in AnyCBM compared to concept interventions in CBM?
    \item Is it possible to train AnyCBM with a dataset slightly different from the one used to train the black-box model?
\end{itemize}
This section describes essential information about the experiments.

\subsection{Data \& task setup}
In our experiments, we use two different datasets commonly used to evaluate CBMs: MNIST even/odd~\citep{barbiero2022entropy}, where the task is to predict whether handwritten digits are even or odd; and CUB~\citep{wah2011caltech}, where the task is to predict bird species based on bird characteristics.

\subsection{Evaluation}
In our analysis, we use ROC-AUC scores to measure classification performance in concepts and downstream tasks and to measure the effectiveness of concept-based interventions in improving classification performance in downstream tasks. To measure the effectiveness of interventions, we follow a similar approach to the one described by ~\citet{zarlenga2022concept}. First, we perturb the latent embeddings by adding a small random noise a few layers before predicting concepts both in AnyCBM and CBM. Then, we intervene on a portion of the concepts with the ground truth. Finally, we test our assumption about the possibility of training AnyCBM with a different dataset with concepts. We train the black-box model with an MNIST even/odd dataset with RGB images. Then, we train AnyCBM with a version of MNIST that contains greyscale images with associated concepts. All results are reported using the mean and standard error over five different runs with different parameter initializations.

\subsection{Baselines}
In our experiments, we compare AnyCBMs with standard CBMs and with an end-to-end black-box model in terms of generalisation performance. We compare AnyCBMs' interventions with the effectiveness of interventions in standard CBMs.

\section{Key findings}
\paragraph{AnyCBMs match black box and CBM performances in terms of classification accuracy on concepts and downstream tasks (Table \ref{tab:task_accuracy}), }
AnyCBMs perform just as well as the original black-box models on which they are based when it comes to accurately completing tasks. Additionally, the accuracy with which these models handle concepts is equal to that of other similar Concept Bottleneck Model architectures. This suggests that AnyCBMs could be a valuable tool for making existing black-box models easier to understand. Using AnyCBMs, we might be able to explain how these complex models work and, in particular, which encoded information is inside the layers of the models, making them more transparent and accessible for further analysis and improvement.

\begin{table}[]
\centering
\begin{tabular}{lcccc}
\hline
 & \multicolumn{2}{c}{\textsc{MNIST even/odd}} & \multicolumn{2}{c}{\textsc{CUB}} \\ 
 & Task ROC AUC & Concept ROC AUC & Task ROC AUC & Concept ROC AUC \\\hline
Black box & $99.8 \pm 0.0$ & - & $90.5 \pm 0.3$ & - \\
CBM & $99.8 \pm 0.0$ & $99.8 \pm 0.0$ & $90.0 \pm 0.2$ & $83.0 \pm 0.2$\\
Black box + AnyCBMs & $99.6 \pm 0.0$ & $98.8 \pm 0.3$ & $90.3 \pm 0.2$ & $84.8 \pm 0.3$ \\ \hline
\end{tabular}%
\caption{Downstream task and concept ROC AUC of AnyCBMs compared to CBMs and a black box model on MNIST and CUB datasets.}
\label{tab:task_accuracy}
\end{table}

\begin{figure}
    \centering
    \begin{minipage}[b]{.45\textwidth}
\centering
\includegraphics[width=\linewidth]{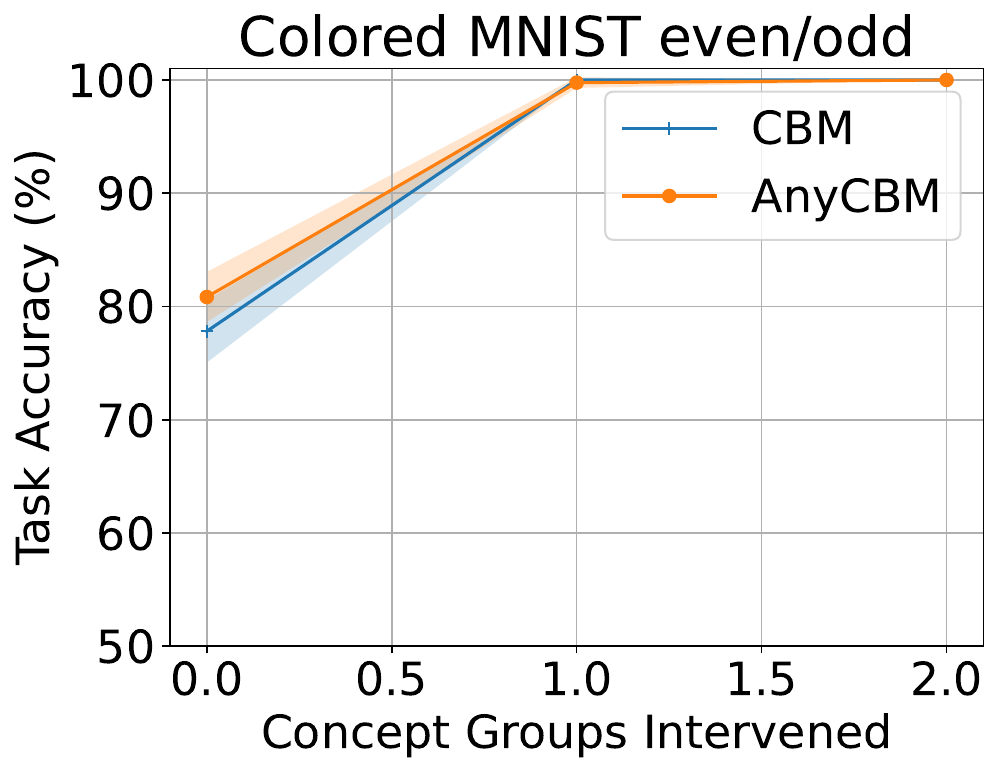}

\end{minipage}
\hfill
\begin{minipage}[b]{.45\textwidth}
\centering
\includegraphics[width=\linewidth]{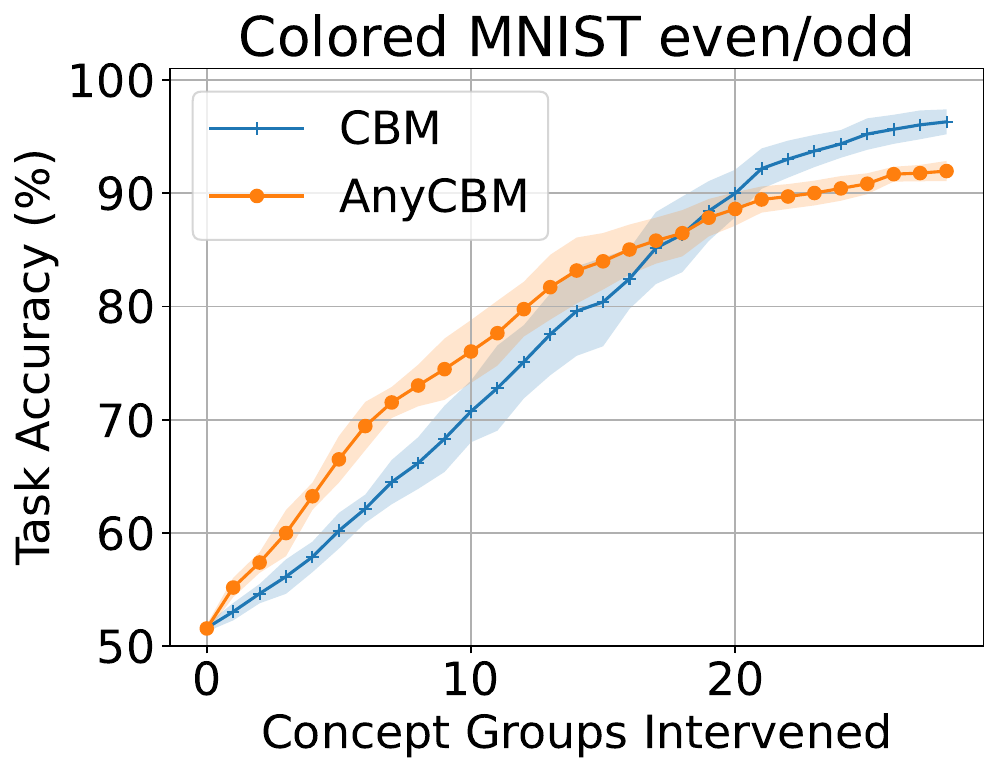}
\end{minipage}\
\caption{Task accuracy of AnyCBMs compared to CBMs after intervening on an increasing number of family of concepts on the MNIST and CUB dataset.}
    \label{fig:interventions}
\end{figure}

\paragraph{AnyCBM interventions are as effective as in Concept Bottleneck Models (Figure \ref{fig:interventions})}
AnyCBMs are as responsive to concept-based interventions as standard CBMs. This means that when concepts predicted by AnyCBMs are manually changed by human experts at test time, they effectively impact the downstream task accuracy. This finding underlines the ability of AnyCBMs to interact with domain experts as it would have been expected by CBMs. In addition, this represents a successful method to steer the behaviour of the model modifying human-understandable concepts. 

\begin{table}[]
\centering
\begin{tabular}{lcccc}
\hline
 & \multicolumn{2}{c}{\textsc{MNIST even/odd RGB}} &\multicolumn{2}{c}{\textsc{MNIST even/odd Grey}}  \\ 
 & Task & Concept & Task & Concept \\\hline
Black box & $98.6 \pm 0.1$ & - & $89.7 \pm 1.7$ & - \\
CBM & $74.4 \pm 3.1$ & $88.7 \pm 0.7$ & $99.3 \pm 0.0$ & $98.6 \pm 0.0$\\
Black box + AnyCBMs & $98.6 \pm 0.0$ & $90.9 \pm 1.2$ & $89.8 \pm 1.6$ & $94.1 \pm 0.2$ \\ \hline
\end{tabular}%
\caption{Downstream task and concept accuracy of AnyCBMs (trained on MNIST Greyscale) compared to CBMs (trained on MNIST Greyscale) and a black box model (trained on MNIST RGB)}
\label{tab:ood}
\end{table}

\paragraph{AnyCBM can be trained with a different dataset from the one used to train the black-box model (Table~\ref{tab:ood})}
One can initially train a black box model with a dataset, which could be larger or more beneficial for addressing the downstream task. Subsequently, the AnyCBM module can be trained on a slightly different dataset that includes concept annotations. As demonstrated in Table~\ref{tab:ood}, this approach does not compromise the model's performance in terms of task accuracy when both the black-box model and AnyCBM are utilised in the original dataset. It also partially accurately predicts concepts in the original dataset, even when there is a distribution shift. This indicates that AnyCBM can alleviate a significant constraint of CBMs, which is the requirement for concept annotations in the dataset used to train the entire model. In addition, the dataset used to train the AnyCBM module could contain only input and concept annotations, without the need for label annotations. 

\section{Discussion}
\paragraph{Advantages}
In the age of Large Models with billions of parameters, the development of solutions that do not require retraining to enhance their capabilities is crucial. AnyCBMs successfully meet this need, as they do not require the alteration of the weights of a pre-trained black-box model. This enables any black-box model to acquire the extra features of CBMs, such as the interpretability of the latent space and the capacity to change the model's behaviour through concept interventions. Furthermore, we believe that AnyCBM can be trained using a dataset that is smaller than the one used to train the original black-box because it has a consistently smaller number of parameters. Interestingly, the dataset can even be distinct (for instance, we might train the model with a dataset without concepts while training AnyCBM with a slightly different dataset that has only concept annotations), mitigating the CBMs' constraint of needing concept annotations for the training set used to train the model. Under these circumstances, it might be intriguing to determine whether certain concepts can be accurately predicted from the latent embeddings of black-box models. If some concepts are unpredictable, this could suggest that the black-box models did not grasp that particular concept in the prior training, either due to the dataset employed or its irrelevant role in task prediction. 
\paragraph{Limitations}
Although the model gains the benefits of CBMs, it also takes on some of their drawbacks. The primary constraint is the necessity for concept data to train the AnyCBM component, although this is somewhat alleviated by the reduced need for concept annotations and the option to utilise an alternate dataset for their extraction.
\paragraph{Future work}
We underscore the importance of delving deeper into AnyCBM and its benefits, while also trying to mitigate its drawbacks. For example, it would be intriguing to examine its application in multimodal contexts, where automatic concept extraction could be feasible, as suggested in \citep{oikarinen2023labelfree}.

\section{Conclusion}
This paper introduces Any Concept Bottleneck Models (AnyCBMs), a method for transforming opaque neural networks into interpretable Concept Bottleneck Models (CBMs), allowing for insights into the decision-making process in terms of concept-based explanations and interventions. This paper analyses practical case studies that demonstrate the properties and limitations of AnyCBMs in enhancing interpretability while maintaining high classification performances from both a theoretical and an experimental perspective. These results suggest how AnyCBMs could represent a computationally effective solution to enhance the interpretability of existing trained or fine-tuned black-box neural networks, allowing also for concept-based interventions in the black-box latent space.

\begin{acknowledgments}
  This study was funded by TRUST-ME (project 205121L\_214991), SmartCHANGE (GA No. 101080965) and XAI-PAC (PZ00P2\_216405). This study was also supported by   TAILOR  project funded by EU Horizon 2020 under GA No 952215. This work has been also supported by the Partnership Extended PE00000013 - “FAIR - Future Artificial Intelligence Research” - Spoke 1 “Human-centered AI”. 
\end{acknowledgments}


\bibliography{sample-ceur}

\end{document}